\documentclass[11pt]{article}

\usepackage[preprint]{acl}

\usepackage{times}
\usepackage{latexsym}

\usepackage[T1]{fontenc}

\usepackage[utf8]{inputenc}

\usepackage{microtype}

\usepackage{inconsolata}

\usepackage{graphicx}

\usepackage{hyperref}
\usepackage{natbib}
\usepackage{multirow}
\usepackage{amsmath}
\usepackage{pifont}
\usepackage{bbm}

\title{GTPred: Benchmarking MLLMs for Interpretable Geo-localization and Time-of-capture Prediction}

\author{First Author \\
  Affiliation / Address line 1 \\
  Affiliation / Address line 2 \\
  Affiliation / Address line 3 \\
  \texttt{email@domain} \\\And
  Second Author \\
  Affiliation / Address line 1 \\
  Affiliation / Address line 2 \\
  Affiliation / Address line 3 \\
  \texttt{email@domain} \\}

\author{
 \textbf{Jinhao Li\textsuperscript{1}},
 \textbf{Tingzhu Chen\textsuperscript{2}},
 \textbf{Changbo Wang\textsuperscript{1}}
\\
 \textsuperscript{1} East China Normal University,
 \textsuperscript{2} Shanghai Jiao Tong University
\\
 \small{
   \textbf{Correspondence:} \href{mailto:email@domain}{lomljhoax@stu.ecnu.edu.cn, tingzhuchen@sjtu.edu.cn, cbwang@cs.ecnu.edu.cn}
 }
}

\begin{document}
\maketitle

\begin{figure*}[t]
    \centering
    \includegraphics[width=\textwidth]{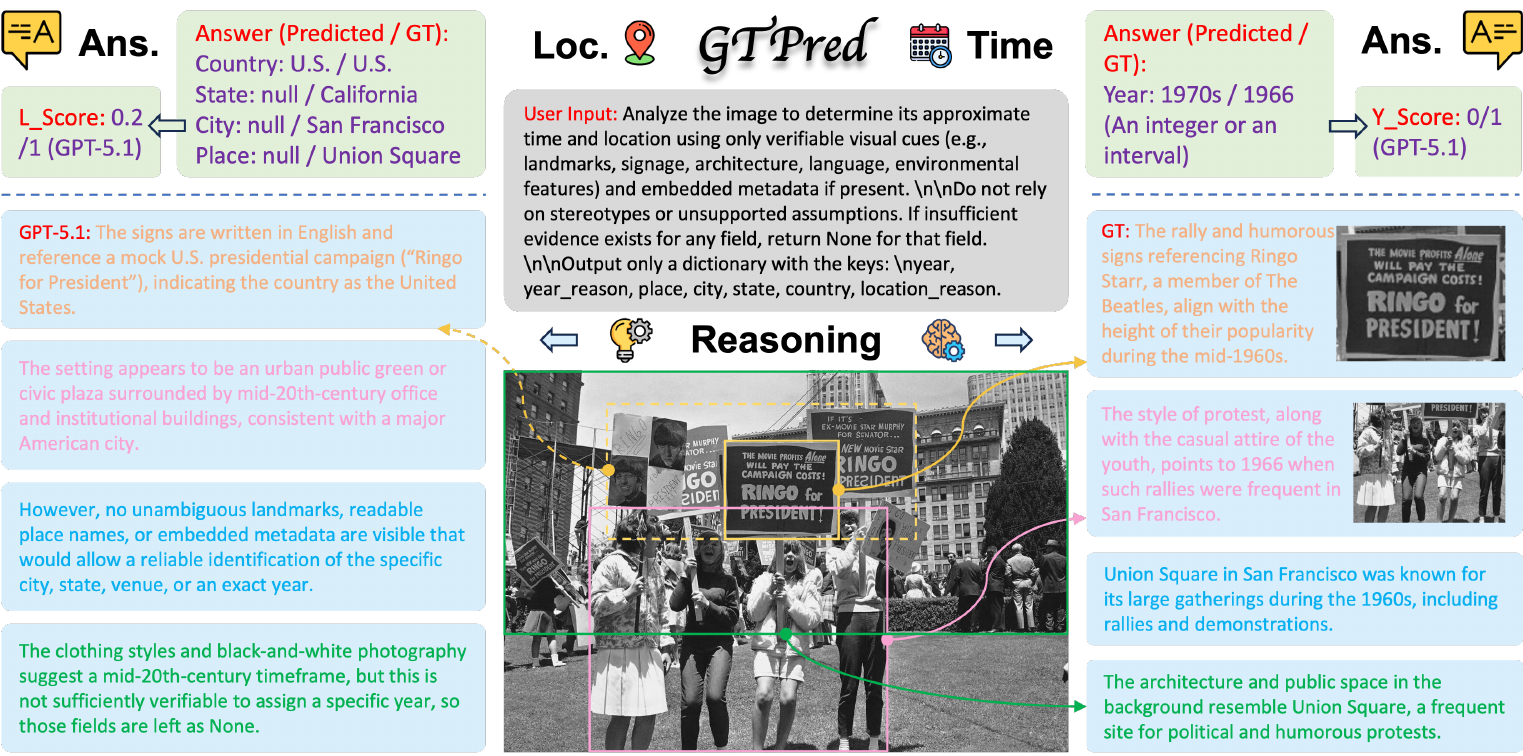}
    \caption{Overview of the proposed GTPred benchmark. Given an image and a prompt, MLLMs are expected to generate both the predicted capture year and location (shaded in green), along with the corresponding reasoning processes (shaded in blue). Particularly, we compare the location reasoning process of GPT-5.1 and the ground truth analysis in detail. The ground truth leverages the predicted year as a cue for reasoning about the location.}
    \label{fig:teaser}
\end{figure*}

\begin{abstract}
Geo-localization aims to infer the geographic location where an image was captured using observable visual evidence. Traditional methods achieve impressive results through large-scale training on massive image corpora. With the emergence of multi-modal large language models (MLLMs), recent studies have explored their applications in geo-localization, benefiting from improved accuracy and interpretability. However, existing benchmarks largely ignore the temporal information inherent in images, which can further constrain geo-localization performance. To bridge this gap, we introduce GTPred, a novel benchmark for geo-temporal prediction. GTPred comprises 370 globally distributed images spanning over 120 years. We evaluate MLLM predictions by jointly considering year and hierarchical location sequence matching, and further assess intermediate reasoning chains using meticulously annotated ground truth reasoning processes. Experiments on 8 proprietary and 7 open-source MLLMs show that, despite strong visual perception, current models remain limited in world knowledge and complex geo-temporal reasoning. Results also demonstrate that incorporating temporal information significantly enhances geo-localization performance. All code, data, and prompts will be made publicly available upon acceptance. 
\end{abstract}

\section{Introduction}

Geo-localization is the task of inferring the geographic location where an image was captured using only observable and verifiable visual cues, such as architectural styles, landmarks, written language, signage, and environmental features. In recent years, the online game GeoGuessr\footnote{https://www.geoguessr.com} has brought increased attention to geo-localization by demonstrating how locations can be inferred from street-level imagery, highlighting both the difficulty and the practical relevance of the problem. Owing to its broad applications in social studies \cite{ye2019visual}, urban planning \cite{shen2017streetvizor}, and navigation \cite{chalvatzaras2022survey}, geo-localization has become an active research topic in computer vision and multi-modal learning. However, the task is inherently challenging due to visual ambiguity, similarities in appearance across geographically distant regions, and the lack of explicit location indicators in many real-world images.

Existing approaches to geo-localization can be broadly categorized into retrieval-based and classification-based methods. Retrieval-based approaches \cite{clark2023we, zhu2022transgeo, lin2022joint, zhang2023cross} identify the most similar image from a geo-tagged image gallery and infer the location accordingly, while classification-based approaches \cite{pramanick2022world, muller2018geolocation, seo2018cplanet, weyand2016planet} discretize the Earth’s surface into predefined regions and assign the input image to one of these regions. With the advent of multi-modal large language models (MLLMs), an increasing number of studies \cite{li2024georeasoner, dou2024gaga, mendes2024granular, yerramilli2025geochain, qian2025earth} have explored their applications in geo-localization, achieving promising performance. Compared to traditional methods, MLLMs benefit from large-scale pretraining and thus possess rich world knowledge that can be leveraged for geo-localization. Moreover, MLLM-based approaches offer improved interpretability by enabling explicit reasoning. However, existing benchmarks \cite{roberts2023gpt4geo, bhandari2023large, huang2025ai, wang2024llmgeo, liu2024image,chen2025just} primarily emphasize visual cues rather than fine-grained reasoning processes for guiding geo-localization. Moreover, they pay limited attention to the temporal information inherent in images, which can provide additional constraints in the inferred location.

As illustrated in Figure~\ref{fig:teaser}, we show the critical role of temporal information in geo-localization through a comparison of GPT-5.1's reasoning process and the ground truth analysis. GPT-5.1 overlooks the temporal cues embedded in the image, ultimately failing to infer the correct location. In contrast, the ground truth reasoning accurately dates the image to 1966 by recognizing references to a member of The Beatles on humorous signs at the rally. Combined with the historical context of frequent public gatherings in Union Square during the 1960s, this temporal evidence enables accurate localization of the image to Union Square, San Francisco, California, United States. While this example compellingly demonstrates the potential of temporal reasoning for geo-localization, it remains unclear whether and how modern MLLMs can reliably perceive, predict, or systematically leverage such temporal information from visual inputs.

Therefore, we aim to benchmark the performance of MLLMs in interpretable geo-temporal prediction. To this end, we introduce GTPred, a geo-temporal prediction benchmark comprising 370 meticulously curated images. Each image is annotated with both the capture year and the geographic location, along with the corresponding reasoning processes. The location annotations follow a hierarchical structure, representing each location as a sequence of levels: Country $\rightarrow$ State $\rightarrow$ City $\rightarrow$ Place. We further propose a novel evaluation protocol within GTPred for assessing the performance of MLLMs. This protocol overcomes the limitations of previous distance-based accuracy by explicitly modeling hierarchical location granularity, and further incorporates temporal information and reasoning processes into a unified evaluation framework. Finally, we conduct extensive experiments on 8 proprietary and 7 open-source MLLMs, providing an in-depth analysis of their geo-temporal prediction performance and highlighting the role of temporal information in geo-localization. In conclusion, the contributions of this paper can be summarized as follows:
\begin{itemize}
    \item We construct GTPred, a novel benchmark for evaluating the geo-temporal prediction capabilities of MLLMs, which contains 370 meticulously curated samples covering a wide range of temporal and geographic contexts.
    \item We propose a novel evaluation protocol that substitutes distance-based metrics with hierarchical location modeling and jointly incorporates temporal information and reasoning processes.
    \item We perform comprehensive evaluations on 8 proprietary and 7 open-source MLLMs, offering an in-depth analysis of their geo-temporal prediction performance and the impact of temporal information on geo-localization.
\end{itemize}

\section{Related Work}

\subsection{Geo-localization}

Research on geo-localization has evolved from early retrieval-based frameworks like IM2GPS \cite{hays2008im2gps} to the geocell-classification paradigm introduced by PlaNet \cite{weyand2016planet}. With the rise of deep learning, Im2GPS3k \cite{vo2017revisiting} established stronger baselines and standardized evaluation splits, while large-scale datasets such as YFCC100M \cite{thomee2016yfcc100m} and the Google landmarks dataset \cite{weyand2020google} enabled global-scale training on massive image corpora. More recently, the field has shifted toward modeling human-level cognitive strategies for geo-localization. For example, PIGEON \cite{haas2024pigeon} leverages GeoGuessr-derived data to evaluate panoramic reasoning. With the emergence of multi-modal large language models (MLLMs), probing their intrinsic geospatial world knowledge has become a central research focus \cite{roberts2023gpt4geo, bhandari2023large, huang2025ai, wang2024llmgeo, liu2024image}. To this end, several recent works \cite{li2024georeasoner, dou2024gaga, mendes2024granular, yerramilli2025geochain, qian2025earth, li2025pixels, talreja2026georc} have introduced large-scale datasets for pretraining and benchmarking, substantially enhancing the visual perception, world knowledge, and complex reasoning capabilities of MLLMs in geo-localization. However, existing benchmarks largely overlook the temporal information associated with the images, which plays an important role in constraining and disambiguating geographic location.

\subsection{Time-of-capture Prediction}

Time-of-capture prediction remains a relatively underexplored problem, with only a limited number of works addressing it directly. Early approaches \cite{tsai2016photo} infer the time of day via sun position and camera geometry, but they rely on strict assumptions, including sky visibility, known GPS coordinates, and access to external databases. Subsequent data-driven methods \cite{zhai2019learning, salem2022timestamp} jointly model time prediction with geo-localization using multi-encoder or hierarchical architectures. However, their performance critically depends on the availability of location information at inference time, which restricts real-world applicability. Most recently, GT-Loc \cite{shatwell2025gt} performs time-of-capture prediction and geo-localization by aligning the image, time, and location embeddings in a shared multi-modal feature space using contrastive learning. Other studies investigate time-of-capture in an indirect or auxiliary manner, for instance, through the verification of claimed timestamps \cite{li2017you, padilha2022content} or the construction of dynamic appearance maps using geo-temporally annotated imagery \cite{salem2020learning}. Besides, works on shadow analysis, sun position estimation, and illumination modeling \cite{jacobs2007geolocating, lalonde2012estimating, wehrwein2015shadow, lalonde2010sun, hold2017deep} provide valuable cues for temporal reasoning. While these methods demonstrate qualitative time estimation capabilities, they likewise depend on known geo-localization or partial temporal metadata, making them unsuitable for fully unconstrained inference.

\section{The GTPred Benchmark}
\label{sec:benchmark}

\subsection{Overview of GTPred}

\begin{table*}[!t]
    \setlength{\tabcolsep}{3pt}
    \centering
    \caption{Comparison of properties between our GTPred and other geo-localization benchmarks}
     \resizebox{1\linewidth}{!}{\begin{tabular}{lcccccc}
	\hline
	  Benchmark  & \# Test Images & Image Sources & Metrics & Verified & Time Rea. & Loc. Rea.  \\ 
    \hline
    IM2GPS \cite{hays2008im2gps} & 237 & Flickr & Acc@k & {\color{red}\ding{55}} & {\color{red}\ding{55}} &  {\color{red}\ding{55}} \\
    YFCC4K \cite{thomee2016yfcc100m} & 4,000 & YFCC100M & Acc@k & {\color{red}\ding{55}} & {\color{red}\ding{55}} & {\color{red}\ding{55}} \\
    LLMGeo \cite{wang2024llmgeo} & 1,000 & GSV & Acc@k & {\color{red}\ding{55}} & {\color{red}\ding{55}} & {\color{red}\ding{55}} \\
    Geo-localizationHub \cite{liu2024image} & 20,000 & GSV & Acc@k, GeoScore & {\color{red}\ding{55}} & {\color{red}\ding{55}} & {\color{red}\ding{55}} \\
    FairLocator \cite{huang2025ai} & 1,200 & GSV & City-level accuracy & {\color{red}\ding{55}} & {\color{red}\ding{55}} & {\color{red}\ding{55}} \\
    GPTGeoChat \cite{mendes2024granular} & 1,000 & Shutterstock & Acc@k & {\color{green}\ding{51}} & {\color{red}\ding{55}} & {\color{red}\ding{55}} \\
    GeoChain \cite{yerramilli2025geochain} & 2,088 & Mapillary & Acc@k, Pass score & {\color{red}\ding{55}} & {\color{red}\ding{55}} & {\color{green}\ding{51}}  \\
    IMAGEO-Bench \cite{li2025pixels} & 9,301 & Public\&Private Data & Acc@k, distance error & {\color{green}\ding{51}} & {\color{red}\ding{55}} & {\color{green}\ding{51}} \\ 
    GeoRC \cite{talreja2026georc} & 500 & GeoGuessr & Precision, Recall, F1, Acc@k & {\color{green}\ding{51}} & {\color{red}\ding{55}} & {\color{green}\ding{51}}  \\
    \hline
    \multirow{2}{*}{WhereBench \cite{qian2025earth}} & \multirow{2}{*}{810} & \multirow{2}{*}{GSV+private} & Acc@k, thinking score & \multirow{2}{*}{{\color{green}\ding{51}}} & \multirow{2}{*}{\color{red}\ding{55}} & \multirow{2}{*}{\color{green}\ding{51}} \\
     & & & hierarchical match & \\
    \hline
    \multirow{2}{*}{GTPred} & \multirow{2}{*}{370} & \multirow{2}{*}{TimeGuessr Explained} & thinking score & \multirow{2}{*}{\color{green}\ding{51}} & \multirow{2}{*}{\color{green}\ding{51}} & \multirow{2}{*}{\color{green}\ding{51}} \\
     & & & hierarchical weighted & \\
    \hline
    \end{tabular}}
    \label{tab:benchmarks}
\end{table*}

\begin{figure*}[t]
    \centering
    \includegraphics[width=\textwidth]{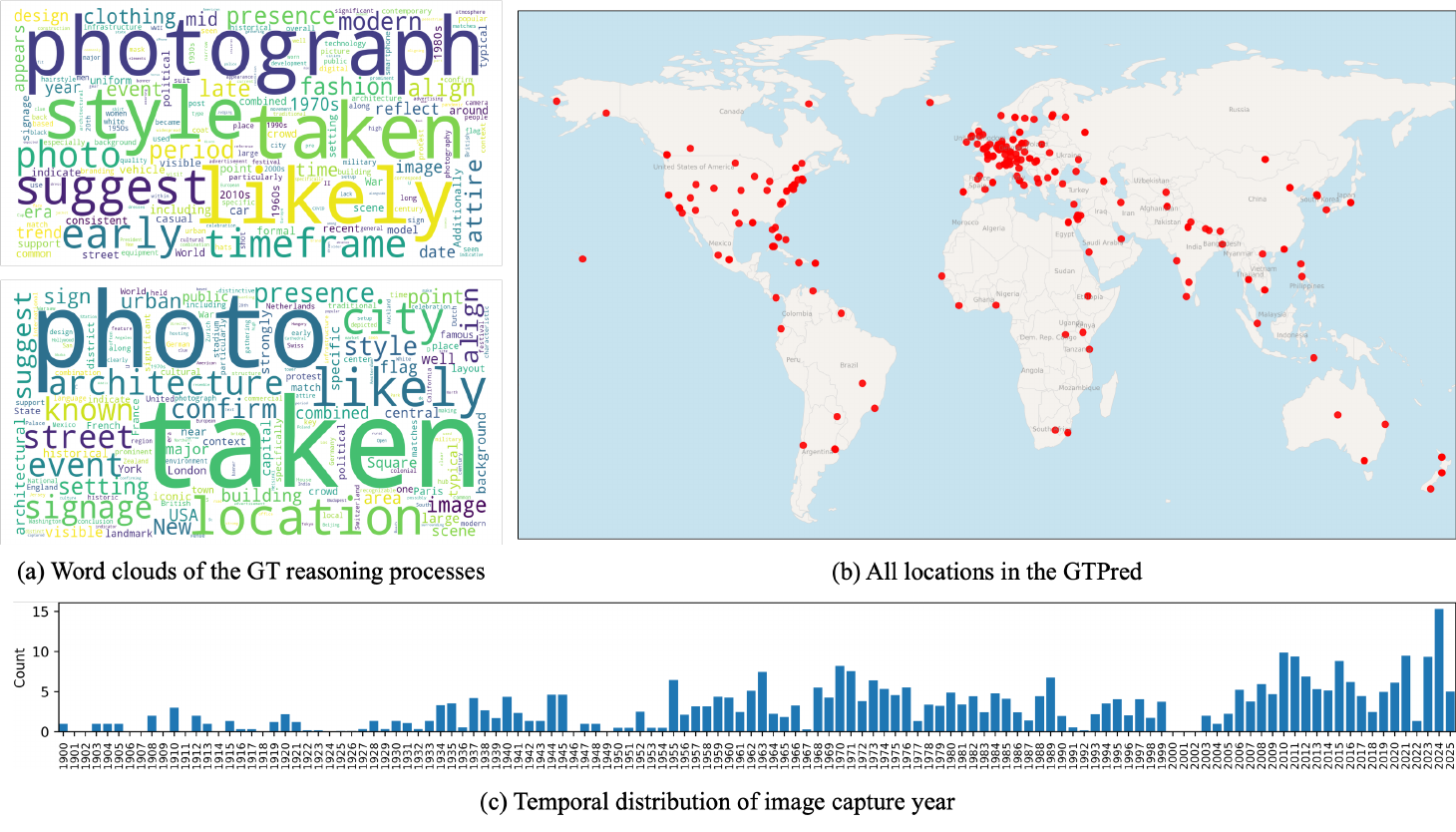}
    \caption{Statistics of the proposed GTPred benchmark. (a) Word clouds of the ground truth reasoning processes for year (upper) and location (lower). (b) All locations in the GTPred shown on a global map. (c) Temporal distribution of image capture year in the GTPred.}
    \label{fig:overview}
\end{figure*}

We introduce GTPred, a novel geo-temporal prediction benchmark designed to comprehensively evaluate the multi-modal understanding capabilities of foundation models in geo-localization and time-of-capture prediction jointly. To our knowledge, GTPred is the first benchmark that incorporates temporal information into the reasoning process to assist multi-modal foundation models in geo-localization. The dataset is meticulously collected from TimeGuessr Explained\footnote{https://www.timeguessrexplained.com/} through a rigorous curation pipeline, encompassing images from 79 countries, 94 states, 186 cities, and 125 fine-grained places across over 120 years. Given an image and a prompt, MLLMs are tasked with jointly predicting the capture year and location while simultaneously generating reasoning processes for both predictions. Crucially, this formulation acknowledges that geographic visual cues are often time-dependent (e.g., evolving fashion or architecture), forcing models to capture dynamic semantic patterns rather than relying on static feature matching. This setup is designed to probe three critical capabilities: visual perception, world knowledge, and complex reasoning. Our primary objective is to assess how effectively these models can not only extract and comprehend multi-modal cues but also apply reasoning with connections between time and location to derive the solution. A detailed comparison of GTPred against existing geo-localization benchmarks is provided in Table~\ref{tab:benchmarks}.

\subsection{Data Curation Pipeline}

The construction of GTPred follows a three-stage pipeline. First, we crawl raw data from TimeGuessr Explained, which publishes daily explanations of the problems in GeoGuessr from June 14, 2024 to March 14, 2025 (with a few missing days). The raw data includes a summary analysis as well as detailed reasoning for both the year and the location. Next, we organize the collected data into a structured JSON format. The year and location answers embedded in the reasoning processes are then automatically extracted using GPT-5.1. To obtain a hierarchical representation of location, we further prompt GPT-5.1 to output a JSON object with four fields: country, state, city, and place. Note that not all samples are annotated with all four levels, which better reflects real-world scenarios where fine-grained location may be partially unavailable. The missing fields are designated as null and are subsequently excluded from downstream evaluation computations. For vague temporal expressions like the early 1990s, mid-1990s, and late 1990s, we adopt a unified standard (e.g., mapping 1990s to the interval 1990–2000) to convert them into well-defined and consistent time intervals. Finally, all samples are manually refined to ensure that the extracted annotations are consistent with the original reasoning processes. As shown in Figure~\ref{fig:overview}, GTPred comprises 370 samples, covering a wide range of temporal and geographic contexts.

\subsection{Evaluation Protocol}

The evaluation metrics in GTPred consider two complementary aspects: the final answer accuracy and the quality of the reasoning process. For time-of-capture prediction, we compute the year score $\mathrm{Y\text{-}score}$ as follows:
\begin{equation}
\mathrm{Y\text{-}score} =
\begin{cases}
\max \left\{0, 1-\frac{|\hat{\mathbf{Y}}-\mathbf{Y}|}{T_{max}}\right\},
\hfill \text{otherwise} \\[4pt]
\mathrm{IoU}(\hat{\mathbf{Y}}, \mathbf{Y}),
\hfill \text{if } \hat{\mathbf{Y}}, \mathbf{Y} \text{ are intervals}
\end{cases}
\end{equation}
where $\hat{\mathbf{Y}}$ and $\mathbf{Y}$ denote the predicted and ground-truth years, respectively, each representing either a single year or an interval. Here, $T_{\max}$ is the maximum year tolerance used for normalization. Note that an integer is equivalent to a degenerate interval containing only one element. 

For geo-localization, the previous studies \cite{vo2017revisiting, weyand2016planet} typically evaluate geo-localization performance using distance-based accuracy at multiple thresholds (e.g., 1–200 km). However, a smaller geographic distance does not necessarily imply a more accurate prediction. For instance, Vatican City and Rome are geographically proximate, yet they are distinct cities belonging to different sovereign entities. To address this issue, we propose a hierarchical weighted score that explicitly accounts for the granularity of correctly matched geographic attributes. The evaluation follows a hierarchical sequence of location levels in both predictions and annotations: Country $\rightarrow$ State $\rightarrow$ City $\rightarrow$ Place. To reflect the varying difficulty across levels, we assign higher rewards to finer-grained matches. Formally, let $\mathbf{L} = [y_1, y_2, y_3, y_4]^{\top}$ denote the ground truth location labels, $\hat{\mathbf{L}} = [\hat{y}_1, \hat{y}_2, \hat{y}_3, \hat{y}_4]^{\top}$ the predicted labels, and $\boldsymbol{\omega} = [0.2, 0.2, 0.3, 0.3]$ the reward vector. The hierarchical weighted score $\mathrm{L\text{-}score}$ is then defined as:
\begin{equation}
    \mathrm{L\text{-}score} = {\rm HWS}(\mathbf{L}, \hat{\mathbf{L}}) = \boldsymbol{\omega} \mathbbm{1}(\mathbf{L}, \hat{\mathbf{L}})
\end{equation}

Beyond final answers, we assess the quality of the model's reasoning process to gain deeper insight into its internal decision-making behavior. For each instance, we construct ground truth reasoning annotations for both the year and location predictions. A strong language model (i.e., GPT-5.1) is then employed as an evaluator, prompted to score the predicted reasoning against the reference on an integer scale from 1 to 10 based on visual evidence, world knowledge, and logical coherence. The raw scores are subsequently linearly normalized to the range [0,1]. This process-level metric captures nuances that answer-only evaluation may overlook, such as valid reasoning that leads to slightly imprecise outputs, or conversely, correct answers derived from flawed logic.
 
\section{Experiments}

\subsection{Experimental Setup}

Our experiments evaluate a total of 15 MLLMs, encompassing a diverse mix of proprietary and open-source models. For proprietary models, we adopt OpenAI models (o4-mini \cite{o3o4}, GPT-4.1 \cite{gpt-4.1}, GPT-5.1 \cite{gpt-5.1}), Gemini models (Gemini 2.5 Flash \cite{comanici2025gemini}, Gemini 2.5 Pro \cite{comanici2025gemini}, and Gemini 3.0 Pro Preview \cite{gemini-3-pro}), and Grok models (Grok-4 \cite{Grok-4} and Grok-4.1 \cite{Grok-4-1}). For open-source models, we cover the Qwen3-VL \cite{Qwen3-VL}, and InternVL3.5 \cite{wang2025internvl3_5} families. We follow all official or recommended inference settings for each MLLM. Reasoning evaluation for GTPred follows the LLM-as-a-Judge framework with GPT-5.1. To avoid information leakage, we conduct all evaluations in a strictly offline setting without Internet access or external retrieval.

\subsection{Comparisons between MLLMs}

\begin{table}[t]
    \setlength{\tabcolsep}{3pt}
    \caption{Comparisons of evaluated MLLMs on geo-localization and time-of-capture prediction. We report the performance on both final answers and reasoning processes. The best and second-best results are in \textbf{bold} and \underline{underlined}, respectively.}
    \label{tab:performance}
    \centering 
    \resizebox{1\linewidth}{!}{\begin{tabular}{lcccc}
    \hline
    \multirow{2}{*}{Models} & \multicolumn{2}{c}{Time} & \multicolumn{2}{c}{Location} \\
    \cline{2-3} \cline{4-5}
     & Answer & Reasoning & Answer & Reasoning \\
    \hline
    \multicolumn{5}{l}{\textbf{Proprietary MLLMs:}} \\
    \hline
    o4-mini & 0.7381 & 0.6288 & 0.7800 & 0.7710 \\
    GPT-4.1 & 0.7384 & 0.7033 & 0.7696 & 0.8022 \\
    GPT-5.1 & 0.7342 & 0.7389 & 0.7670 & 0.8301 \\
    Gemini 2.5 Flash & 0.7725 & 0.7375 & 0.7873 & 0.8463 \\
    Gemini 2.5 Pro & \underline{0.8554} & \underline{0.7953} & \underline{0.8228} & \underline{0.8682} \\
    Gemini 3 Pro & \textbf{0.8762} & \textbf{0.8121} & \textbf{0.8526} & \textbf{0.8707} \\
    Grok-4 & 0.5914 & 0.6630 & 0.7019 & 0.7805 \\
    Grok-4.1 & 0.6132 & 0.6929 & 0.7177 & 0.8003 \\
    \hline
    \multicolumn{5}{l}{\textbf{Open-source MLLMs:}} \\
    \hline
    Qwen3-VL-4B & 0.3063 & 0.4915 & 0.5680 & 0.6756 \\
    Qwen3-VL-8B & 0.3907 & 0.5485 & 0.6214 & 0.7225 \\
    Qwen3-VL-32B & 0.5884 & 0.7485 & 0.7003 & 0.7625 \\
    InternVL3.5-4B & 0.0780 & 0.2963 & 0.2222 & 0.4973 \\
    InternVL3.5-8B & 0.2481 & 0.4479 & 0.3782 & 0.4729 \\
    InternVL3.5-14B & 0.0460 & 0.2951 & 0.2284 & 0.4652 \\
    InternVL3.5-38B & 0.1762 & 0.3888 & 0.3673 & 0.5129 \\
    \hline
    \end{tabular}}
\end{table}

As summarized in Table~\ref{tab:performance}, proprietary MLLMs demonstrate superior performance in both geo-localization and time-of-capture prediction. In terms of answer accuracy, the leading proprietary model achieves scores of 0.8526 for location and 0.8762 for time prediction, significantly outperforming the open-source alternatives. Even the strongest open-source model reaches only 0.7003 (location) and 0.5884 (year), lagging behind by margins of 0.1523 and 0.2878, respectively. This performance gap is further reflected in the reasoning-based metrics, where proprietary models consistently generate higher-quality reasoning processes. Notably, the performance gap reaches 0.2878 in time-of-capture prediction, substantially exceeding the 0.1523 margin observed in geo-localization, suggesting that temporal-contextual alignment remains a key weakness of open-source MLLMs. Given the strong positive correlation between answer accuracy and reasoning quality across models, we define their product as the composite score to holistically evaluate performance in the subsequent experiments.

\subsection{Scaling Behavior}

\begin{figure}[t]
    \centering
    \includegraphics[width=\linewidth]{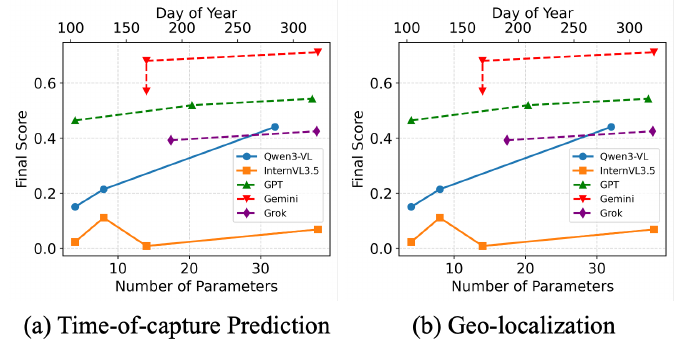}
    \caption{Scaling behavior of different model families on our GTPred.}
    \label{fig:scale}
\end{figure}

To exemplify how MLLMs benefit from scaling laws in geo-localization and time-of-capture prediction, we compare their performance and visualize the results in Figure~\ref{fig:scale}. For proprietary MLLMs, performance is reported w.r.t. their release dates, quantified as the day of the year in 2025. The overall averaged performance adheres to the expected scaling trend, underscoring the potential of scaling MLLMs for improving geo-temporal prediction performance. Nevertheless, we also observe that larger models within the same family do not always outperform their smaller counterparts. For instance, InternVL3.5-38B fails to achieve competitive performance despite its larger scale. These findings suggest that scaling alone is insufficient and that effective geo-temporal reasoning also relies on data quality, multi-modal alignment, and the ability to exploit fine-grained visual and temporal cues.

\subsection{Ablation Study}

\begin{table}[t]
    \caption{Ablation study on the role of temporal and geographic cues in geo-localization and time-of-capture prediction.}
    \label{tab:ablation}
    \centering 
    \resizebox{1\linewidth}{!}{\begin{tabular}{lcccc}
    \hline
    \multirow{2}{*}{Models} & \multicolumn{2}{c}{Time} & \multicolumn{2}{c}{Location} \\
    \cline{2-3} \cline{4-5}
     & w/ loc. & w/o loc. & w/ time & w/o time \\
    \hline
    \multicolumn{5}{l}{\textbf{Proprietary MLLMs:}} \\
    \hline
    o4-mini & 0.4641 & 0.5145 & 0.6014 & 0.5870 \\
    GPT-4.1 & 0.5193 & 0.5597 & 0.6174 & 0.5405 \\
    GPT-5.1 & 0.5425 & 0.5174 & 0.6367 & 0.6126 \\
    \hline
    \multicolumn{5}{l}{\textbf{Open-source MLLMs:}} \\
    \hline
    Qwen3-VL-4B & 0.1505 & 0.0340 & 0.3837 & 0.3975 \\
    Qwen3-VL-8B & 0.2143 & 0.3606 & 0.4490 & 0.3849 \\
    Qwen3-VL-32B & 0.4404 & 0.5551 & 0.5340 & 0.5133 \\
    \hline
    \end{tabular}}
\end{table}

\noindent \textbf{Impact of Temporal and Geographic Information.} We investigate the impact of temporal and geographic information on MLLM's performance by separately evaluating geo-localization and time-of-capture prediction. The results are summarized in Table~\ref{tab:ablation}. We observe that most MLLMs suffer a substantial performance degradation in geo-localization when temporal information is removed. For instance, Qwen3-VL-8B shows a performance drop of 0.0641, while GPT-4.1 exhibits an even larger decline of 0.0769. However, geographic information contributes little to time-of-capture prediction, and in some cases, removing geographic cues even leads to improved time-of-capture prediction performance. We assume that this phenomenon arises because geographic information provides only weak or non-discriminative constraints on time-of-capture prediction. Overall, these results reveal an asymmetric contribution of temporal and geographic information in geo-temporal prediction: temporal cues are critical for geo-localization, whereas geographic cues provide limited benefit for time-of-capture prediction.

\begin{table*}[t]
    \caption{Results of the reasoning evaluation using GPT-5.1, Grok-4.1 and Qwen3-VL-32B as judges. The best and second-best results are in \textbf{bold} and \underline{underlined}, respectively.}
    \label{tab:bias}
    \centering 
    \resizebox{1\linewidth}{!}{\begin{tabular}{lccccccc}
    \hline
    \multirow{2}{*}{Models} & \multicolumn{3}{c}{Time} & \multicolumn{3}{c}{Location} \\
    \cline{2-4} \cline{5-7}
     & GPT-5.1 & Grok-4.1 & Qwen3-VL-32B & GPT-5.1 & Grok-4.1 & Qwen3-VL-32B \\
    \hline
    \textbf{Proprietary MLLMs:} & \\
    \hline
    o4-mini & 0.4641 & 0.5277 & 0.6033 & 0.6014 & 0.7153 & 0.7099 \\
    GPT-4.1 & 0.5193 & 0.6077 & 0.6726 & 0.6174 & 0.7329 & 0.7449 \\
    GPT-5.1 & 0.5425 & 0.5866 & 0.6901 & 0.6367 & 0.7567 & \underline{0.7644} \\
    Gemini 2.5 Flash & 0.5697 & 0.6458 & 0.6786 & 0.6663 & 0.7827 & 0.7581 \\
    Gemini 2.5 Pro & \underline{0.6803} & \underline{0.6989} & \underline{0.7101} & \underline{0.7144} & \underline{0.8036} & \underline{0.7644} \\
    Gemini 3 Pro & \textbf{0.7116} & \textbf{0.7219} & \textbf{0.7132} & \textbf{0.7424} & \textbf{0.8184} & \textbf{0.7666} \\
    Grok-4 & 0.3921 & 0.5419 & 0.6188 & 0.5478 & 0.7014 & 0.7044 \\
    Grok-4.1 & 0.4249 & 0.5734 & 0.6268 & 0.5744 & 0.7332 & 0.7244 \\
    \hline
    \textbf{Open-source MLLMs:} & \\
    \hline
    Qwen3-VL-4B & 0.1505 & 0.1100 & 0.1496 & 0.3837 & 0.3606 & 0.3357 \\
    Qwen3-VL-8B & 0.2143 & 0.2275 & 0.1956 & 0.4490 & 0.4189 & 0.3857 \\
    Qwen3-VL-32B & 0.4404 & 0.3661 & 0.3763 & 0.5340 & 0.4886 & 0.4652 \\
    InternVL3.5-4B & 0.0231 & 0.0230 & 0.0084 & 0.1105 & 0.1415 & 0.1494 \\
    InternVL3.5-8B & 0.1111 & 0.0222 & 0.0172 & 0.1789 & 0.1598 & 0.1532 \\
    InternVL3.5-14B & 0.0136 & 0.0184 & 0.0077 & 0.1063 & 0.1334 & 0.1339 \\
    InternVL3.5-38B & 0.0685 & 0.0790 & 0.0635 & 0.1884 & 0.2703 & 0.2340 \\
    \hline
    \end{tabular}}
\end{table*}

\noindent \textbf{Validation of Different Judges.} We adopt GPT-5.1 as the judge to evaluate the reasoning processes of MLLMs in the preceding experiments. To mitigate potential self-preference bias and ensure an objective assessment, we additionally employ Grok-4.1 and Qwen3-VL-32B as independent judges, thereby enhancing the robustness and reliability of the LLM-as-a-Judge evaluation. All judges follow the same evaluation prompt with clearly defined criteria covering visual evidence, world knowledge, and logical coherence. As shown in Figure~\ref{tab:bias}, although the absolute scores vary across different judges, consistent conclusions can be drawn regarding performance differences both across model families and among models with different parameter scales within the same family. This strong consensus across diverse judges underscores the validity and generalizability of our LLM-as-a-Judge evaluation framework.

\subsection{Human Evaluation}

To further validate the reliability of the LLM-as-a-Judge framework, we conduct a human evaluation. The participants consist of 15 master’s or Ph.D. students in computer science who had no prior exposure to our benchmark dataset. They are tasked with rating the MLLM-generated reasoning processes against the ground truth annotations on a 10-point Likert scale, focusing on visual evidence, world knowledge, and logical coherence. To alleviate evaluation fatigue and ensure high-quality annotation, we randomly sampled 100 test instances and collected human ratings for the reasoning outputs of GPT-5.1, Gemini 3 Pro Preview, and Grok-4.1. Crucially, the user study was conducted in a double-blind fashion, where all model identities were anonymized, and the presentation order was strictly randomized. The results are summarized in Table~\ref{tab:human}. It can be observed that the results of the human evaluation align with those of the LLM-as-a-Judge evaluation.

\begin{table}[t]
    \setlength{\tabcolsep}{3pt}
    \caption{Human evaluation results of GPT-5.1, Gemini 3 Pro Preview, and Grok-4.1. We also report the results of LLM-as-a-Judge (GPT-5.1 as judge) for comparisons.}
    \label{tab:human}
    \centering 
    \resizebox{1\linewidth}{!}{\begin{tabular}{lcccc}
    \hline
    \multirow{2}{*}{Models} & \multicolumn{2}{c}{Time Reasoning} & \multicolumn{2}{c}{Location Reasoning} \\
    \cline{2-3} \cline{4-5}
     & Automatic & Human & Automatic & Human \\
    \hline
    GPT-5.1 & 0.7389 & 0.6951 & 0.8301 & 0.7230 \\
    Gemini 3 Pro & 0.8121 & 0.7863 & 0.8707 & 0.8019 \\
    Grok-4.1 & 0.6929 & 0.5630 & 0.8003 & 0.6704 \\
    \hline
    \end{tabular}}
\end{table}

\subsection{Case Study}

\begin{figure}[t]
    \centering
    \includegraphics[width=\linewidth]{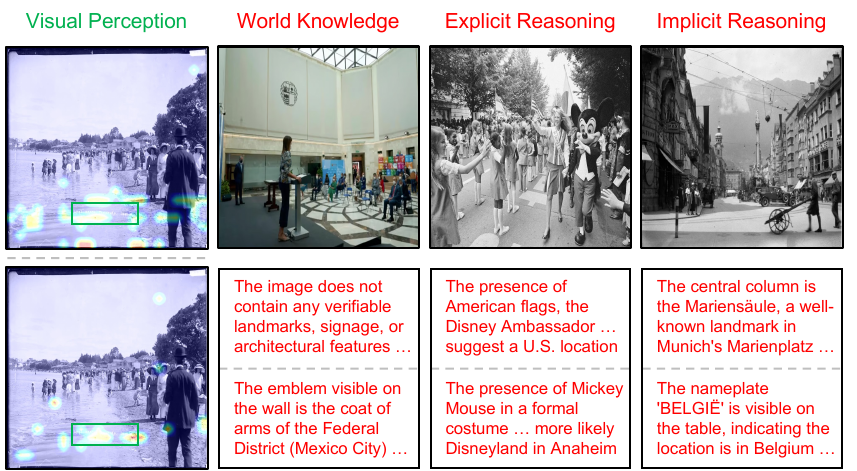}
    \caption{Examples of successful and failed cases from Qwen3-VL-4B and Qwen3-VL-32B. The results from them are separated by dashed lines. The green box indicates the signage in the image.}
    \label{fig:vis}
\end{figure}

As shown in Figure~\ref{fig:vis}, we present representative examples spanning four types of challenges: visual perception, world knowledge, explicit reasoning, and implicit reasoning. In the first and relatively straightforward case (Column 1), the Qwen3-VL family successfully extracts location information from blurred handwritten signage, demonstrating robust visual perception even under degraded visual conditions. In contrast, the second case reveals a limitation in fine-grained world knowledge. Qwen3-VL-32B erroneously identified the emblem on the wall as the coat of arms of Mexico City, whereas it actually belongs to the Basque Country, Spain. The remaining cases involve reasoning challenges, which can be further divided into explicit and implicit reasoning. In the third case (Column 3), the image contains explicit geographical cues in the form of the United States and Japanese flags. However, the model fails to recognize that the depicted event corresponds to Disney parades, which were commonly used in cultural promotions during the early 1970s in Japan, indicating limitations in explicit reasoning. The fourth case (Column 4) represents a more demanding implicit reasoning scenario. Successful localization requires synthesizing multiple visual and historical cues (architecture, language, environmental features) into a coherent, multi-step inference. The model is unable to complete this reasoning process, resulting in a clear localization failure. Overall, these examples demonstrate that while MLLMs exhibit strong visual perception capabilities, their performance deteriorates substantially when accurate localization depends on fine-grained world knowledge or multi-step complex reasoning.

\section{Discussion} 

As elaborated in Section~\ref{sec:benchmark}, the data from TimeGuessr Explained was collected via web crawling and published between June 14, 2024 and March 14, 2025. While web-scraped data inherently carries a risk of contamination in model training corpora, our experimental evidence suggests that GTPred does not suffer from meaningful data leakage. Specifically, this publication window falls after the training cutoff of o4-mini and GPT-4.1 but before those of the Qwen3-VL and InternVL3.5 families. If data leakage were present, models with later training cutoffs (i.e., Qwen3-VL and InternVL3.5 families) would be expected to outperform o4-mini and GPT-4.1. However, the results demonstrate the opposite trend. Furthermore, this period also precedes the training cutoff of GPT-5.1, yet GPT-5.1 does not exhibit a substantial performance advantage over o4-mini and GPT-4.1. Therefore, these observations indicate that GTPred is unlikely to have been incorporated into the training sets of the evaluated models.

\section{Conclusion}

In this paper, we introduce GTPred, a novel benchmark specifically designed for geo-temporal prediction. The GTPred consists of 370 meticulously annotated samples spanning a comprehensive range of temporal and geographic contexts. We further propose a process-aware evaluation protocol that integrates year intervals and hierarchical location sequence matching with an LLM-as-a-Judge rubric to comprehensively evaluate the performance of MLLMs on geo-temporal prediction. Extensive experiments demonstrate that while MLLMs excel at visual perception, they exhibit limited capabilities in world knowledge and complex reasoning. Moreover, our findings highlight the effectiveness of temporal information in facilitating geographic inference. We hope these results can provide insights into the future improvements of MLLMs for geo-localization and time-of-capture prediction.

\section{Limitations}

One limitation of GTPred is its relatively modest scale compared to some large-scale geo-localization benchmarks. Although its size is comparable to or larger than several widely used benchmarks such as WhereBench and IM2GPS, the current dataset may still limit the diversity of geographic regions, temporal distributions, and rare edge cases covered in evaluation. In addition, the human annotation and verification process, while improving label quality and reasoning reliability, also constrains the speed of dataset expansion. Nevertheless, the underlying data source, TimeGuessr, is continuously updated and contains a substantially larger pool of candidate samples. Future work should further expand GTPred in both scale and diversity to improve benchmark coverage and support more comprehensive evaluation. Furthermore, because GTPred jointly evaluates geo-localization and time-of-capture prediction, the task is inherently more challenging and may introduce compounded uncertainty when spatial or temporal cues are ambiguous.



\begin{thebibliography}{50}
\providecommand{\natexlab}[1]{#1}

\bibitem[{Bai et~al.(2025)}]{Qwen3-VL}
Shuai Bai and 1 others. 2025.
\newblock Qwen3-vl technical report.
\newblock \emph{arXiv preprint arXiv:2511.21631}.

\bibitem[{Bhandari et~al.(2023)Bhandari, Anastasopoulos, and Pfoser}]{bhandari2023large}
Prabin Bhandari, Antonios Anastasopoulos, and Dieter Pfoser. 2023.
\newblock Are large language models geospatially knowledgeable?
\newblock In \emph{Proceedings of the 31st ACM International Conference on Advances in Geographic Information Systems}, pages 1--4.

\bibitem[{Chalvatzaras et~al.(2022)Chalvatzaras, Pratikakis, and Amanatiadis}]{chalvatzaras2022survey}
Athanasios Chalvatzaras, Ioannis Pratikakis, and Angelos~A Amanatiadis. 2022.
\newblock A survey on map-based localization techniques for autonomous vehicles.
\newblock \emph{IEEE Transactions on intelligent vehicles}, 8(2):1574--1596.

\bibitem[{Chen et~al.(2025)Chen, Tian, Sun, Sun, Zhang, Lin, Zhai, and Zhang}]{chen2025just}
Zijian Chen, Yuan Tian, Yuze Sun, Wei Sun, Zicheng Zhang, Weisi Lin, Guangtao Zhai, and Wenjun Zhang. 2025.
\newblock Just noticeable difference for large multimodal models.
\newblock \emph{arXiv preprint arXiv:2507.00490}.

\bibitem[{Clark et~al.(2023)Clark, Kerrigan, Kulkarni, Cepeda, and Shah}]{clark2023we}
Brandon Clark, Alec Kerrigan, Parth~Parag Kulkarni, Vicente~Vivanco Cepeda, and Mubarak Shah. 2023.
\newblock Where we are and what we're looking at: Query based worldwide image geo-localization using hierarchies and scenes.
\newblock In \emph{Proceedings of the IEEE/CVF Conference on Computer Vision and Pattern Recognition}, pages 23182--23190.

\bibitem[{Comanici et~al.(2025)Comanici, Bieber, Schaekermann, Pasupat, Sachdeva, Dhillon, Blistein, Ram, Zhang, Rosen et~al.}]{comanici2025gemini}
Gheorghe Comanici, Eric Bieber, Mike Schaekermann, Ice Pasupat, Noveen Sachdeva, Inderjit Dhillon, Marcel Blistein, Ori Ram, Dan Zhang, Evan Rosen, and 1 others. 2025.
\newblock Gemini 2.5: Pushing the frontier with advanced reasoning, multimodality, long context, and next generation agentic capabilities.
\newblock \emph{arXiv preprint arXiv:2507.06261}.

\bibitem[{Deepmind(2025)}]{gemini-3-pro}
Google Deepmind. 2025.
\newblock Gemini 3 pro.
\newblock \url{https://deepmind.google/mode-ls/gemini/pro/}.
\newblock Accessed: Nov 19, 2025.

\bibitem[{Dou et~al.(2024)Dou, Wang, Han, Li, Huang, and Han}]{dou2024gaga}
Zhiyang Dou, Zipeng Wang, Xumeng Han, Guorong Li, Zhipei Huang, and Zhenjun Han. 2024.
\newblock Gaga: Towards interactive global geolocation assistant.
\newblock \emph{arXiv preprint arXiv:2412.08907}.

\bibitem[{Haas et~al.(2024)Haas, Skreta, Alberti, and Finn}]{haas2024pigeon}
Lukas Haas, Michal Skreta, Silas Alberti, and Chelsea Finn. 2024.
\newblock Pigeon: Predicting image geolocations.
\newblock In \emph{Proceedings of the IEEE/CVF Conference on Computer Vision and Pattern Recognition}, pages 12893--12902.

\bibitem[{Hays and Efros(2008)}]{hays2008im2gps}
James Hays and Alexei~A Efros. 2008.
\newblock Im2gps: estimating geographic information from a single image.
\newblock In \emph{2008 ieee conference on computer vision and pattern recognition}, pages 1--8. IEEE.

\bibitem[{Hold-Geoffroy et~al.(2017)Hold-Geoffroy, Sunkavalli, Hadap, Gambaretto, and Lalonde}]{hold2017deep}
Yannick Hold-Geoffroy, Kalyan Sunkavalli, Sunil Hadap, Emiliano Gambaretto, and Jean-Fran{\c{c}}ois Lalonde. 2017.
\newblock Deep outdoor illumination estimation.
\newblock In \emph{Proceedings of the IEEE conference on computer vision and pattern recognition}, pages 7312--7321.

\bibitem[{Huang et~al.(2025)Huang, Huang, Liu, Liu, Wang, and Zhao}]{huang2025ai}
Jingyuan Huang, Jen-tse Huang, Ziyi Liu, Xiaoyuan Liu, Wenxuan Wang, and Jieyu Zhao. 2025.
\newblock Ai sees your location—but with a bias toward the wealthy world.
\newblock In \emph{Proceedings of the 2025 Conference on Empirical Methods in Natural Language Processing}, pages 18030--18050.

\bibitem[{Jacobs et~al.(2007)Jacobs, Satkin, Roman, Speyer, and Pless}]{jacobs2007geolocating}
Nathan Jacobs, Scott Satkin, Nathaniel Roman, Richard Speyer, and Robert Pless. 2007.
\newblock Geolocating static cameras.
\newblock In \emph{2007 IEEE 11th International Conference on Computer Vision}, pages 1--6. IEEE.

\bibitem[{Lalonde et~al.(2012)Lalonde, Efros, and Narasimhan}]{lalonde2012estimating}
Jean-Fran{\c{c}}ois Lalonde, Alexei~A Efros, and Srinivasa~G Narasimhan. 2012.
\newblock Estimating the natural illumination conditions from a single outdoor image.
\newblock \emph{International Journal of Computer Vision}, 98(2):123--145.

\bibitem[{Lalonde et~al.(2010)Lalonde, Narasimhan, and Efros}]{lalonde2010sun}
Jean-Fran{\c{c}}ois Lalonde, Srinivasa~G Narasimhan, and Alexei~A Efros. 2010.
\newblock What do the sun and the sky tell us about the camera?
\newblock \emph{International Journal of Computer Vision}, 88(1):24--51.

\bibitem[{Li et~al.(2024)Li, Ye, Zhou, Jiang, and Zeng}]{li2024georeasoner}
Ling Li, Yu~Ye, Yao Zhou, Bingchuan Jiang, and Wei Zeng. 2024.
\newblock Georeasoner: Geo-localization with reasoning in street views using a large vision-language model.
\newblock \emph{arXiv preprint arXiv:2406.18572}.

\bibitem[{Li et~al.(2025)Li, Yu, Hu, Li, Deng, Zhou, and Jia}]{li2025pixels}
Lingyao Li, Runlong Yu, Qikai Hu, Bowei Li, Min Deng, Yang Zhou, and Xiaowei Jia. 2025.
\newblock From pixels to places: A systematic benchmark for evaluating image geolocalization ability in large language models.
\newblock \emph{arXiv preprint arXiv:2508.01608}.

\bibitem[{Li et~al.(2017)Li, Xu, Wang, and Qu}]{li2017you}
Xiaopeng Li, Wenyuan Xu, Song Wang, and Xianshan Qu. 2017.
\newblock Are you lying: Validating the time-location of outdoor images.
\newblock In \emph{International Conference on Applied Cryptography and Network Security}, pages 103--123. Springer.

\bibitem[{Lin et~al.(2022)Lin, Zheng, Zhong, Luo, Li, Yang, and Sebe}]{lin2022joint}
Jinliang Lin, Zhedong Zheng, Zhun Zhong, Zhiming Luo, Shaozi Li, Yi~Yang, and Nicu Sebe. 2022.
\newblock Joint representation learning and keypoint detection for cross-view geo-localization.
\newblock \emph{IEEE Transactions on Image Processing}, 31:3780--3792.

\bibitem[{Liu et~al.(2024)Liu, Ding, Deng, Li, Zhang, Sun, Zheng, Ge, and Liu}]{liu2024image}
Yi~Liu, Junchen Ding, Gelei Deng, Yuekang Li, Tianwei Zhang, Weisong Sun, Yaowen Zheng, Jingquan Ge, and Yang Liu. 2024.
\newblock Image-based geolocation using large vision-language models.
\newblock \emph{arXiv preprint arXiv:2408.09474}.

\bibitem[{Mendes et~al.(2024)Mendes, Chen, Hays, Das, Xu, and Ritter}]{mendes2024granular}
Ethan Mendes, Yang Chen, James Hays, Sauvik Das, Wei Xu, and Alan Ritter. 2024.
\newblock Granular privacy control for geolocation with vision language models.
\newblock \emph{arXiv preprint arXiv:2407.04952}.

\bibitem[{Muller-Budack et~al.(2018)Muller-Budack, Pustu-Iren, and Ewerth}]{muller2018geolocation}
Eric Muller-Budack, Kader Pustu-Iren, and Ralph Ewerth. 2018.
\newblock Geolocation estimation of photos using a hierarchical model and scene classification.
\newblock In \emph{Proceedings of the European conference on computer vision (ECCV)}, pages 563--579.

\bibitem[{OpenAI(2025{\natexlab{a}})}]{gpt-4.1}
OpenAI. 2025{\natexlab{a}}.
\newblock Gpt-4.1.
\newblock \url{https://openai.com/index/gpt-4-1}.
\newblock Accessed: Apr 14, 2025.

\bibitem[{OpenAI(2025{\natexlab{b}})}]{gpt-5.1}
OpenAI. 2025{\natexlab{b}}.
\newblock Gpt-5.1.
\newblock \url{https://openai.com/index/gpt-5-1/}.
\newblock Accessed: Nov 12, 2025.

\bibitem[{OpenAI(2025{\natexlab{c}})}]{o3o4}
OpenAI. 2025{\natexlab{c}}.
\newblock o3 and o4-mini.
\newblock \url{https://openai.com/index/introducing-o3-and-o4-mini/}.
\newblock Accessed: Apr 16, 2025.

\bibitem[{Padilha et~al.(2022)Padilha, Salem, Workman, Andal{\'o}, Rocha, and Jacobs}]{padilha2022content}
Rafael Padilha, Tawfiq Salem, Scott Workman, Fernanda~A Andal{\'o}, Anderson Rocha, and Nathan Jacobs. 2022.
\newblock Content-aware detection of temporal metadata manipulation.
\newblock \emph{IEEE Transactions on Information Forensics and Security}, 17:1316--1327.

\bibitem[{Pramanick et~al.(2022)Pramanick, Nowara, Gleason, Castillo, and Chellappa}]{pramanick2022world}
Shraman Pramanick, Ewa~M Nowara, Joshua Gleason, Carlos~D Castillo, and Rama Chellappa. 2022.
\newblock Where in the world is this image? transformer-based geo-localization in the wild.
\newblock In \emph{European Conference on Computer Vision}, pages 196--215. Springer.

\bibitem[{Qian et~al.(2025)Qian, Chen, Wang, Zhang, Wang, Huang, Liu, Tang, Zheng, Tu et~al.}]{qian2025earth}
Zhaofang Qian, Hardy Chen, Zeyu Wang, Li~Zhang, Zijun Wang, Xiaoke Huang, Hui Liu, Xianfeng Tang, Zeyu Zheng, Haoqin Tu, and 1 others. 2025.
\newblock Where on earth? a vision-language benchmark for probing model geolocation skills across scales.
\newblock \emph{arXiv preprint arXiv:2510.10880}.

\bibitem[{Roberts et~al.(2023)Roberts, L{\"u}ddecke, Das, Han, and Albanie}]{roberts2023gpt4geo}
Jonathan Roberts, Timo L{\"u}ddecke, Sowmen Das, Kai Han, and Samuel Albanie. 2023.
\newblock Gpt4geo: How a language model sees the world's geography.
\newblock \emph{arXiv preprint arXiv:2306.00020}.

\bibitem[{Salem et~al.(2022)Salem, Hwang, and Padilha}]{salem2022timestamp}
Tawfiq Salem, Jisoo Hwang, and Rafael Padilha. 2022.
\newblock Timestamp estimation from outdoor scenes.

\bibitem[{Salem et~al.(2020)Salem, Workman, and Jacobs}]{salem2020learning}
Tawfiq Salem, Scott Workman, and Nathan Jacobs. 2020.
\newblock Learning a dynamic map of visual appearance.
\newblock In \emph{Proceedings of the IEEE/CVF Conference on Computer Vision and Pattern Recognition}, pages 12435--12444.

\bibitem[{Seo et~al.(2018)Seo, Weyand, Sim, and Han}]{seo2018cplanet}
Paul~Hongsuck Seo, Tobias Weyand, Jack Sim, and Bohyung Han. 2018.
\newblock Cplanet: Enhancing image geolocalization by combinatorial partitioning of maps.
\newblock In \emph{Proceedings of the European Conference on Computer Vision (ECCV)}, pages 536--551.

\bibitem[{Shatwell et~al.(2025)Shatwell, Dave, Swetha, and Shah}]{shatwell2025gt}
David~G Shatwell, Ishan~Rajendrakumar Dave, Sirnam Swetha, and Mubarak Shah. 2025.
\newblock Gt-loc: Unifying when and where in images through a joint embedding space.
\newblock In \emph{Proceedings of the IEEE/CVF International Conference on Computer Vision}, pages 1--11.

\bibitem[{Shen et~al.(2017)Shen, Zeng, Ye, Arisona, Schubiger, Burkhard, and Qu}]{shen2017streetvizor}
Qiaomu Shen, Wei Zeng, Yu~Ye, Stefan~M{\"u}ller Arisona, Simon Schubiger, Remo Burkhard, and Huamin Qu. 2017.
\newblock Streetvizor: Visual exploration of human-scale urban forms based on street views.
\newblock \emph{IEEE transactions on visualization and computer graphics}, 24(1):1004--1013.

\bibitem[{Talreja et~al.(2026)Talreja, Diao, James, Casapu, Santanam, Mendes, Ritter, Xu, and Hays}]{talreja2026georc}
Mohit Talreja, Joshua Diao, Jim~Thannikary James, Radu Casapu, Tejas Santanam, Ethan Mendes, Alan Ritter, Wei Xu, and James Hays. 2026.
\newblock Georc: A benchmark for geolocation reasoning chains.
\newblock \emph{arXiv preprint arXiv:2601.21278}.

\bibitem[{Thomee et~al.(2016)Thomee, Shamma, Friedland, Elizalde, Ni, Poland, Borth, and Li}]{thomee2016yfcc100m}
Bart Thomee, David~A Shamma, Gerald Friedland, Benjamin Elizalde, Karl Ni, Douglas Poland, Damian Borth, and Li-Jia Li. 2016.
\newblock Yfcc100m: The new data in multimedia research.
\newblock \emph{Communications of the ACM}, 59(2):64--73.

\bibitem[{Tsai et~al.(2016)Tsai, Jhou, Cheng, Hu, Shen, Lim, Hua, Ghoneim, Hossain, and Hidayati}]{tsai2016photo}
Tsung-Hung Tsai, Wei-Cih Jhou, Wen-Huang Cheng, Min-Chun Hu, I-Chao Shen, Tekoing Lim, Kai-Lung Hua, Ahmed Ghoneim, M~Anwar Hossain, and Shintami~C Hidayati. 2016.
\newblock Photo sundial: estimating the time of capture in consumer photos.
\newblock \emph{Neurocomputing}, 177:529--542.

\bibitem[{Vo et~al.(2017)Vo, Jacobs, and Hays}]{vo2017revisiting}
Nam Vo, Nathan Jacobs, and James Hays. 2017.
\newblock Revisiting im2gps in the deep learning era.
\newblock In \emph{Proceedings of the IEEE international conference on computer vision}, pages 2621--2630.

\bibitem[{Wang et~al.(2025)Wang, Gao, Gu, Pu, Cui, Wei, Liu, Jing, Ye, Shao et~al.}]{wang2025internvl3_5}
Weiyun Wang, Zhangwei Gao, Lixin Gu, Hengjun Pu, Long Cui, Xingguang Wei, Zhaoyang Liu, Linglin Jing, Shenglong Ye, Jie Shao, and 1 others. 2025.
\newblock Internvl3.5: Advancing open-source multimodal models in versatility, reasoning, and efficiency.
\newblock \emph{arXiv preprint arXiv:2508.18265}.

\bibitem[{Wang et~al.(2024)Wang, Xu, Khan, Lin, Fan, and Zhu}]{wang2024llmgeo}
Zhiqiang Wang, Dejia Xu, Rana Muhammad~Shahroz Khan, Yanbin Lin, Zhiwen Fan, and Xingquan Zhu. 2024.
\newblock Llmgeo: Benchmarking large language models on image geolocation in-the-wild.
\newblock \emph{arXiv preprint arXiv:2405.20363}.

\bibitem[{Wehrwein et~al.(2015)Wehrwein, Bala, and Snavely}]{wehrwein2015shadow}
Scott Wehrwein, Kavita Bala, and Noah Snavely. 2015.
\newblock Shadow detection and sun direction in photo collections.
\newblock In \emph{2015 International Conference on 3D Vision}, pages 460--468. IEEE.

\bibitem[{Weyand et~al.(2020)Weyand, Araujo, Cao, and Sim}]{weyand2020google}
Tobias Weyand, Andre Araujo, Bingyi Cao, and Jack Sim. 2020.
\newblock Google landmarks dataset v2-a large-scale benchmark for instance-level recognition and retrieval.
\newblock In \emph{Proceedings of the IEEE/CVF conference on computer vision and pattern recognition}, pages 2575--2584.

\bibitem[{Weyand et~al.(2016)Weyand, Kostrikov, and Philbin}]{weyand2016planet}
Tobias Weyand, Ilya Kostrikov, and James Philbin. 2016.
\newblock Planet-photo geolocation with convolutional neural networks.
\newblock In \emph{European conference on computer vision}, pages 37--55. Springer.

\bibitem[{xAI(2025{\natexlab{a}})}]{Grok-4}
xAI. 2025{\natexlab{a}}.
\newblock Grok 4.
\newblock \url{https://x.ai/news/grok-4}.
\newblock Accessed: Aug 1, 2025.

\bibitem[{xAI(2025{\natexlab{b}})}]{Grok-4-1}
xAI. 2025{\natexlab{b}}.
\newblock Grok 4.1.
\newblock \url{https://x.ai/news/grok-4-1}.
\newblock Accessed: Nov 17, 2025.

\bibitem[{Ye et~al.(2019)Ye, Zeng, Shen, Zhang, and Lu}]{ye2019visual}
Yu~Ye, Wei Zeng, Qiaomu Shen, Xiaohu Zhang, and Yi~Lu. 2019.
\newblock The visual quality of streets: A human-centred continuous measurement based on machine learning algorithms and street view images.
\newblock \emph{Environment and Planning B: Urban Analytics and City Science}, 46(8):1439--1457.

\bibitem[{Yerramilli et~al.(2025)Yerramilli, Pande, Grover, and Tamarapalli}]{yerramilli2025geochain}
Sahiti Yerramilli, Nilay Pande, Rynaa Grover, and Jayant~Sravan Tamarapalli. 2025.
\newblock Geochain: Multimodal chain-of-thought for geographic reasoning.
\newblock \emph{arXiv preprint arXiv:2506.00785}.

\bibitem[{Zhai et~al.(2019)Zhai, Salem, Greenwell, Workman, Pless, and Jacobs}]{zhai2019learning}
Menghua Zhai, Tawfiq Salem, Connor Greenwell, Scott Workman, Robert Pless, and Nathan Jacobs. 2019.
\newblock Learning geo-temporal image features.
\newblock \emph{arXiv preprint arXiv:1909.07499}.

\bibitem[{Zhang et~al.(2023)Zhang, Li, Sultani, Zhou, and Wshah}]{zhang2023cross}
Xiaohan Zhang, Xingyu Li, Waqas Sultani, Yi~Zhou, and Safwan Wshah. 2023.
\newblock Cross-view geo-localization via learning disentangled geometric layout correspondence.
\newblock In \emph{Proceedings of the AAAI conference on artificial intelligence}, volume~37, pages 3480--3488.

\bibitem[{Zhu et~al.(2022)Zhu, Shah, and Chen}]{zhu2022transgeo}
Sijie Zhu, Mubarak Shah, and Chen Chen. 2022.
\newblock Transgeo: Transformer is all you need for cross-view image geo-localization.
\newblock In \emph{Proceedings of the IEEE/CVF Conference on Computer Vision and Pattern Recognition}, pages 1162--1171.

\end{thebibliography}




\end{document}